\documentclass[sigconf]{acmart}
\usepackage{pgfplots}

\AtBeginDocument{%
  }

\acmISBN{978-1-4503-XXXX-X/2018/06}

\usepackage{amsmath,amssymb,amsfonts}

\usepackage{graphicx}
\usepackage{footnote}
\usepackage{textcomp}
\usepackage{xcolor}
\usepackage{algpseudocode}
\usepackage{algorithm}
\usepackage{graphicx}   % for \rotatebox
\usepackage{multirow}   % for \multirow
\usepackage[dvipsnames]{xcolor}
\usepackage{pgfplots}
\pgfplotsset{compat=newest}
\usepgfplotslibrary{groupplots}
\usepackage{geometry}
\usepackage{graphicx}
\usepackage{siunitx}
\usepackage{hyperref}
\usepackage{xcolor}
\usepackage{subfig}
\usepackage{comment}
\usepackage{pgfplots}
\usepackage{pgfplotstable}

\usetikzlibrary{matrix,backgrounds}

\usetikzlibrary{patterns}

\newboolean{showcomments}
 \setboolean{showcomments}{true}
\usepackage{sepfootnotes}
 \setboolean{showcomments}{false}
\ifthenelse{\boolean{showcomments}}
{ \newcommand{\mynote}[3]{
    \fbox{\bfseries\sffamily\scriptsize#1}
    {\small$\blacktriangleright$\textsf{\emph{\color{#3}{#2}}}$\blacktriangleleft$}}}
{ \newcommand{\mynote}[3]{}}
\newcommand{\shrink}[1]{}
\usepackage{ifthen} % in the preamble

\ifthenelse{\boolean{showcomments}}{
    \newcommand{\new}[1]{\textcolor{teal}{#1}}
    \newcommand{\mb}[1]{\textcolor{orange}{\textbf{ Meriem:} #1}}
    \newcommand{\jb}[1]{\textcolor{blue}{\textbf{ Jalil:} #1}}
     \newcommand{\yh}[1]{\textcolor{violet}{\textbf{ Yuan-Hao} #1}}
    
    }
{
    \newcommand{\new}[1]{}
    \newcommand{\mb}[1]{}
    \newcommand{\jb}[1]{}
    \newcommand{\yh}[1]{}

}
\setcopyright{none}
\renewcommand\footnotetextcopyrightpermission[1]{}
\acmConference{}{}{}{}
\acmBooktitle{}
\acmDOI{}
\acmISBN{}
\begin{document}
\fancyhead{}

%%
%% The "title" command has an optional parameter,
%% allowing the author to define a "short title" to be used in page headers.
%\title{APreQEL: Adaptive Mixed Precision Quantization For Edge LLMs: Trading off memory footprint, accuracy and latency}
\title{APreQEL: Adaptive Mixed Precision Quantization
For Edge LLMs}
%%
%% The "author" command and its associated commands are used to define
%% the authors and their affiliations.
%% Of note is the shared affiliation of the first two authors, and the
%% "authornote" and "authornotemark" commands
%% used to denote shared contribution to the research.
% \author{Ben Trovato}
% \authornote{Both authors contributed equally to this research.}
% \email{trovato@corporation.com}
% \orcid{1234-5678-9012}
% \author{G.K.M. Tobin}
% \authornotemark[1]
% \email{webmaster@marysville-ohio.com}
% \affiliation{%
%   \institution{Institute for Clarity in Documentation}
%   \city{Dublin}
%   \state{Ohio}
%   \country{USA}
% }

\author{Meriem Bouzouad}
\affiliation{%
  \institution{Lab-STICC, CNRS UMR 6285 , ENSTA, Institut Polytechnique de Paris}
  \city{Brest}
  \country{France}}
\email{meriem.bouzouad@ensta.fr}

\author{Yuan-Hao Chang}
\affiliation{%
  \institution{National Taiwan University}
  \city{Taipei}
  \country{Taiwan}}
\email{johnson@csie.ntu.edu.tw}

\author{Jalil Boukhobza}
\affiliation{%
  \institution{Lab-STICC, CNRS UMR 6285 , ENSTA, Institut Polytechnique de Paris}
  \city{Brest}
  \country{France}}
\email{jalil.boukhobza@ensta.fr}

% \author{Valerie B\'eranger}
% \affiliation{%
%   \institution{Inria Paris-Rocquencourt}
%   \city{Rocquencourt}
%   \country{France}
% }

%%
%% By default, the full list of authors will be used in the page
%% headers. Often, this list is too long, and will overlap
%% other information printed in the page headers. This command allows
%% the author to define a more concise list
%% of authors' names for this purpose.
\renewcommand{\shortauthors}{Trovato et al.}

%%
%% The abstract is a short summary of the work to be presented in the
%% article.
\begin{abstract}
Today, large language models have demonstrated their strengths in various tasks ranging from reasoning, code generation, and complex problem solving. However, this advancement comes with a high computational cost and memory requirements, making it challenging to deploy these models on edge devices to ensure real-time responses and data privacy.
Quantization is one common approach to reducing memory use, but most methods apply it uniformly across all layers. This does not account for the fact that different layers may respond differently to reduced precision. 
Importantly, memory consumption and computational throughput are not necessarily aligned, further complicating deployment decisions.
This paper proposes an adaptive mixed precision quantization mechanism that balances memory, latency, and accuracy in edge deployment under user-defined priorities. This is achieved by analyzing the layer-wise contribution and by inferring how different quantization types behave across the target hardware platform in order to assign the most suitable quantization type to each layer. This integration ensures that layer importance and the overall performance trade-offs are jointly respected in this design. Our work unlocks new configuration designs that uniform quantization cannot achieve, expanding the solution space to efficiently deploy the LLMs on resource-constrained devices.
\end{abstract}

%%
%% The code below is generated by the tool at http://dl.acm.org/ccs.cfm.
%% Please copy and paste the code instead of the example below.
%%
% \begin{CCSXML}
% <ccs2012>
%  <concept>
%   <concept_id>00000000.0000000.0000000</concept_id>
%   <concept_desc>Do Not Use This Code, Generate the Correct Terms for Your Paper</concept_desc>
%   <concept_significance>500</concept_significance>
%  </concept>
%  <concept>
%   <concept_id>00000000.00000000.00000000</concept_id>
%   <concept_desc>Do Not Use This Code, Generate the Correct Terms for Your Paper</concept_desc>
%   <concept_significance>300</concept_significance>
%  </concept>
%  <concept>
%   <concept_id>00000000.00000000.00000000</concept_id>
%   <concept_desc>Do Not Use This Code, Generate the Correct Terms for Your Paper</concept_desc>
%   <concept_significance>100</concept_significance>
%  </concept>
%  <concept>
%   <concept_id>00000000.00000000.00000000</concept_id>
%   <concept_desc>Do Not Use This Code, Generate the Correct Terms for Your Paper</concept_desc>
%   <concept_significance>100</concept_significance>
%  </concept>
% </ccs2012>
% \end{CCSXML}

% \ccsdesc[500]{Do Not Use This Code~Generate the Correct Terms for Your Paper}
% \ccsdesc[300]{Do Not Use This Code~Generate the Correct Terms for Your Paper}
% \ccsdesc{Do Not Use This Code~Generate the Correct Terms for Your Paper}
% \ccsdesc[100]{Do Not Use This Code~Generate the Correct Terms for Your Paper}

%%
%% Keywords. The author(s) should pick words that accurately describe
%% the work being presented. Separate the keywords with commas.
%% A "teaser" image appears between the author and affiliation
%% information and the body of the document, and typically spans the
%% page.

% \received{20 February 2007}
% \received[revised]{12 March 2009}
% \received[accepted]{5 June 2009}

%%
%% This command processes the author and affiliation and title
%% information and builds the first part of the formatted document.

\maketitle

\section{Introduction}

 The enhancement of reasoning capabilities in large language models (LLMs) is mainly associated with the growth in their number of parameters. Thus, enabling those models to capture complex relationships and dependencies from data \cite{wei2022emergent}. However, such scale-up necessitates substantial computational power and significant memory capacity, often requiring deployment on resource-rich cloud platforms for inference requests, especially to ensure scalability and assure quality of service for the users. This shift raises various security concerns, as well as a communication overhead, particularly due to internet connection and availability issues in certain regions.

With the advent of edge intelligence and real-time processing of data, deploying LLM models on the edge has gained much interest in research areas~\cite{WANG2025100755}, especially with advancements in hardware accelerators in edge devices, such as integrated GPU, NPU, and powerful CPUs. However, the memory bandwidth does not scale with the compute speed, which hinders the overall performance of systems, particularly for LLMs, which require a significant amount of memory to load the weight matrices and context requests. Indeed, deploying LLMs on the edge requires having a reasonable \textbf{Quality of Service} metrics in terms of \textbf{memory footprint} while ensuring good \textbf{accuracy} with limited \textbf{latency}.

One of the most adopted strategies to deploy those models efficiently is quantization, which reduces the precision of weights, generally from 16-bit to 8-bit or even 2-bit. This has allowed those large models to run on GPU consumer hardware or even on CPU; however, the hardware support for operations under 8-bits is limited to a few types of AI accelerators, which are rarely present in edge devices. As a result, the acceleration is not always reachable, since the de-quantization overhead can offset the theoretical speedup expected from memory footprint reduction.

Many quantization schemes have appeared in recent years such as GPTQ~\cite{frantar2022gptq}, AWQ~\cite{lin2024awq}, SmoothQuant\cite{pmlr-v202-xiao23c}, K-quant types from llama.cpp\footnote{https://github.com/ggml-org/llama.cpp} library, which algorithm is formulated in \cite{egashira2025mind}. 
Some recent work, such as ShortGPT\cite{men-etal-2025-shortgpt}, have exploited the similarity between layers to prune the layers judged unimportant,  while others \cite{dumitru-etal-2025-variable} assign variable quantization depending on the importance of the layers; however, in addition to being limited to one QoS metric objective (memory footprint), their experiments are limited to 2-bit and 4-bit quantization. In addition, except for \cite{dumitru-etal-2025-variable}, all the techniques cited above follow a uniform quantization scheme, which treats all layers in an LLM equally. 
However, not every layer affects accuracy in the same way. 
Some layers are critical and require higher precision, while others can tolerate lower precision without significant loss of accuracy \cite{dumitru-etal-2025-variable}; uniform quantization ignores this difference, often causing unnecessary reductions in accuracy or missing opportunities for more efficient compression\cite{10.1145/3649329.3658473}.

 Moreover, uniform quantization may lead to a degraded QoS, in terms of inference latency and memory if hardware-specific characteristics are ignored. This is because different quantization methods introduce varying, often uncorrelated impacts on latency and memory depending on the model architecture, as reported in the motivation Section~(\ref{motiv:inf-lat}), largely due to de-quantization overhead. As a result, model compression may fail to translate into actual efficiency gains on a given target hardware.

To the best of our knowledge, previous studies did not explore the trade-offs achievable through mixed quantization schemes, particularly when latency is included. 
This limitation is particularly problematic in edge environments, where resource availability often fluctuates more than in the cloud, requiring designs that are both adaptable and efficient to meet shifting requirements.

In this paper, we propose APreQEL, a flexible solution for layer-wise quantization in LLMs, which adapts to diverse deployment scenarios by incorporating user-driven QoS requirements. APreQEL makes it possible for users to make a trade-off between QoS metrics (memory footprint, model accuracy, and latency). 

To this end, our approach (1)~ranks layers depending on their contribution to the change of information; %explicitly differentiating between attention layers and FFN layers;
(2)~determines the best distribution of quantization types, accounting for their distinct QoS metrics;
(3)~combines the ranking from step (1) with the distribution from step (2) to assign the most suitable quantization type to each layer.

Our findings demonstrate that APreQEL systematically guarantees that the generated solutions are spread over the Pareto-front, under any priority set by the user, enabling either focused optimization on a single metric, joint optimization of two metrics, or balanced trade-offs across memory footprint, accuracy, and latency.
Notably, APreQEL achieves $\sim$9\% increase in the hypervolume indicator on average over the uniform baseline set, reflecting a broader Pareto front while considering only a small set of 28 configurations.

\section{Background and Motivation}
\subsection{Background}
\label{background}
 LLMs are built on top of a decoder-only transformer architecture \cite{chen2024decoder}. The natural language prompt is converted to tokens, which are mapped to high-dimensional vectors called embeddings to capture their semantic content. Those embeddings are then passed through $N$ decoding blocks, each containing a self-attention layer and a feedforward network (FNN) layer, to capture the contextual relationship between the different embeddings. Finally, LM\_head module predicts the next token.

\subsection{Motivation1: Layers contribute differently}
In this part, we examine how the intermediate representation (embeddings) evolves across transformer layers to assess whether layers contribute differently, suggesting that distinct layers may benefit from different quantization strategies.
To conduct this observation, we used the Llama3.1 model, where token-level cosine similarity is computed using 70 prompts from the huggingface WebInstructSub-prometheus dataset~\footnote{https://huggingface.co/datasets/chargoddard/WebInstructSub-prometheus} and the transformers library of python, with 32bit precision.
\begin{figure}
    \centering
    \includegraphics[width=1\linewidth]{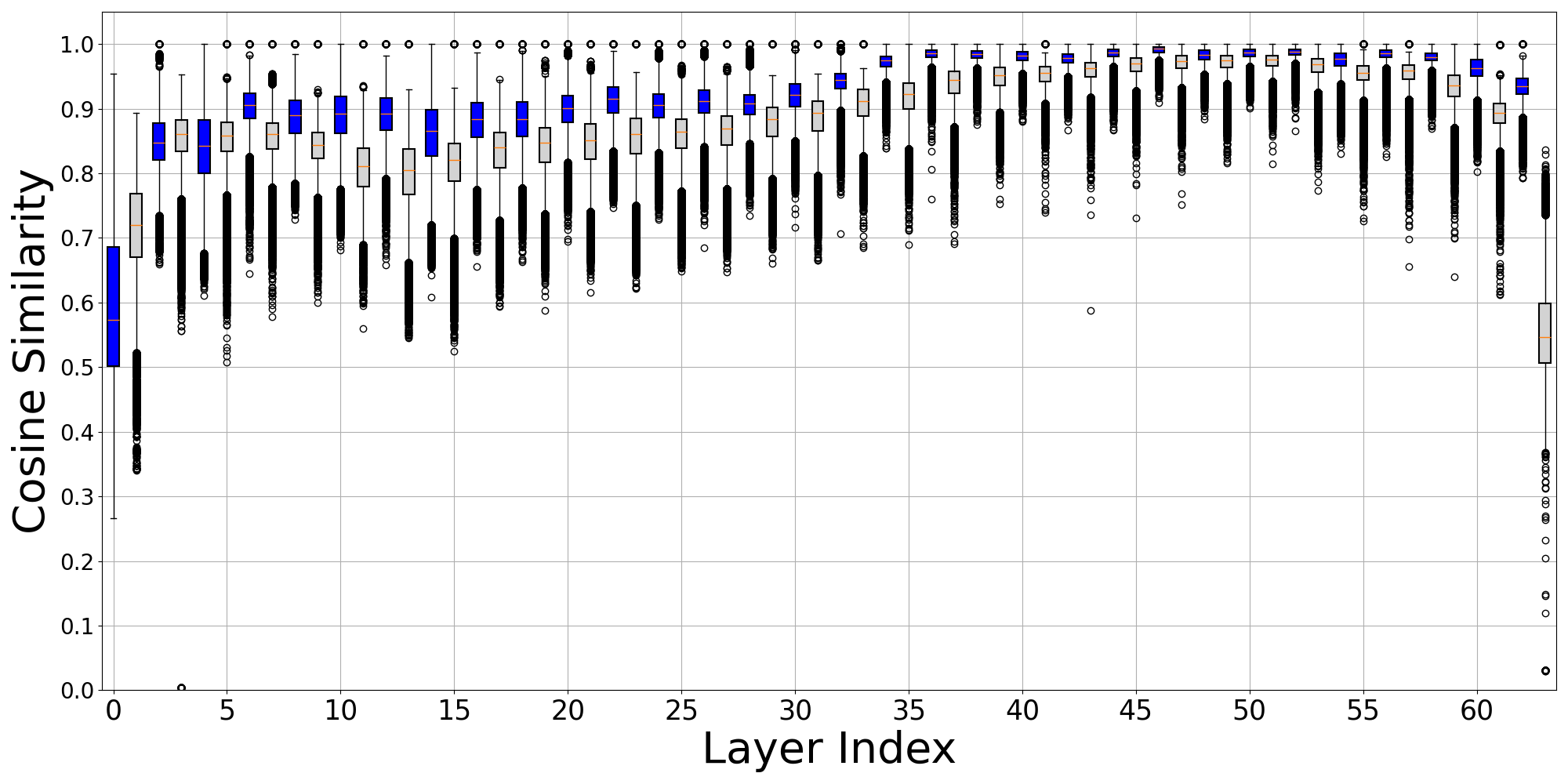}
    \caption{Cosine similarity distribution for Llama-3.1\protect\footnotemark~including Attention and FFN layers.}
    \label{fig:cosine-sim-ffn-att}
\end{figure}
     \footnotetext{https://huggingface.co/meta-llama/Llama-3.1-8B-Instruct} 

Figure~\ref{fig:cosine-sim-ffn-att} illustrates the cosine similarity of hidden embeddings across consecutive layers, accounting for both the attention (blue) and the feed-forward network (grey). We observe  different trends;  sometimes attention layers are closer to 1 (highly similar), and sometimes the feed-forward network layer is, depending on their position; This leads us to score them separately, which provides a better choice for quantization, unlike \cite{dumitru-etal-2025-variable} that collapsed them inside the decoder block.%~(see Section~\ref{background}).

\subsection{Motivation 2: Unpredictable Inference Latency across quantization types}\label{motiv:inf-lat}

We have examined how latency varies depending on both the type of quantization and the specific model architecture.
We used the popular K-types supported by llama.cpp for 3, 4, 5, and 6-bit quantization, and q8\_0 for 8-bit; We employed the \textit{llama-bench} tool to measure the time required to generate a single token over the hardware setup in Section \ref{hardware-setup}, with the GPU enabled.

\begin{figure}[tbhp]
\centering
\resizebox{0.9\linewidth}{!}{
% First plot
\begin{minipage}{0.3\textwidth}
\centering
\begin{tikzpicture}
\begin{axis}[
    ybar,
    bar width=7pt,
    width=5cm,
    height=5.5cm,
    enlarge x limits=0.000025,
    ymin=0, ymax=160,
    xmin=1.2, xmax=4.2,
    ylabel={Latency~(ms)},
    xlabel={ Model size~(GB)~(Phi3.5)},
    ylabel style={yshift=-5pt},
    xlabel style={
      % /pgf/number format/use comma,   % decimal comma
      % /pgf/number format/fixed,
      /pgf/number format/precision=1  % one digit after comma
    },
 %   symbolic x coords={1.53, 2.0, 2.45, 2.9, 3},
    % xtick=data,
    nodes near coords,
    point meta=explicit symbolic,
    every node near coord/.append style={font=\footnotesize, yshift=1pt},
]
\addplot+[fill=blue!50] coordinates {
   (1.53,133.5) [3-bit]
 (2.0,76.6) [4-bit]
 (2.45,82.7) [5-bit]
 (2.92,108.7) [6-bit]
 (3.78,106.3) [8-bit]
};
\end{axis}
\end{tikzpicture}
%\caption{PHI3.5 model}
% \caption[margin={2cm,0cm}]{PHI3.5}

\label{fig:modelA}
\end{minipage}
\hfill
% Second plot
\begin{minipage}{0.3\textwidth}
\hspace{-1.1cm}
\centering
\begin{tikzpicture}
\begin{axis}[
    ybar,
    bar width=7pt,
    width=5cm,
    height=5.5cm,
    enlarge x limits=0.000025,
    ymin=0, ymax=280,
    xmin=2.5, xmax=9,
    xlabel={ Model size~(GB)~(Llama3.1)},
    xlabel style={
      /pgf/number format/use comma,   % decimal comma
      % /pgf/number format/fixed,
      /pgf/number format/precision=1  % one digit after comma
    },
    nodes near coords,
    point meta=explicit symbolic,
    every node near coord/.append style={font=\footnotesize, yshift=1pt},
]
\addplot+[fill=blue!50] coordinates {
(3.21,232.0) [3-bit]
 (4.21,123.9) [4-bit]
 (5.14,138.7) [5-bit]
 (6.14,191.2) [6-bit]
 (8,133.3) [8-bit]
};
\end{axis}
\end{tikzpicture}

%\caption[margin={2cm,0cm}]{Llama3.1}
\label{fig:modelB}
\end{minipage}
}
% Global caption
\caption{Inference Latency by model size for Phi3.5 model (left) and Llama3.1 model (right) according to the quantization levels}
\label{fig:side-by-side-plots}
\end{figure}
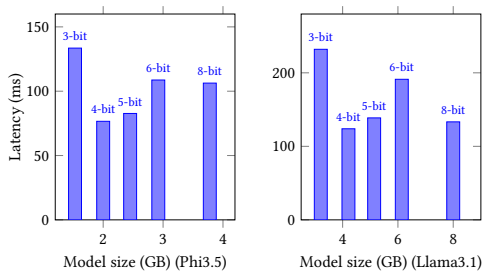
Figure \ref{fig:side-by-side-plots} illustrates the (stable) average time required to generate one token over 5 runs for different quantization types in llama.cpp during the text generation phase, revealing the lack of a clear correlation between model size and inference time. For example, we see that 5-bit quantization is faster than 8-bit for the Phi3.5 model, contrary to the Llama3.1 model. Our main hypothesis is that this behavior stems from de-quantization overhead, which occurs when converting the quantized weights to the original float format for the computational operations. This is due to the lack of hardware support, the algorithm used for de-quantization, and the model architecture.
This suggests that aggressive quantization techniques, while effective in reducing model size, can lead to either gains or losses in inference latency, depending heavily on both the model architecture and the underlying hardware.

 Our objective, therefore, is to explore mixed-quantization strategies that can leverage the strengths of each uniform type in order to adapt to diverse deployment needs in terms of accuracy, latency, and memory.

\section{Contribution}

\begin{figure}
    \centering
    \includegraphics[width=1\linewidth]{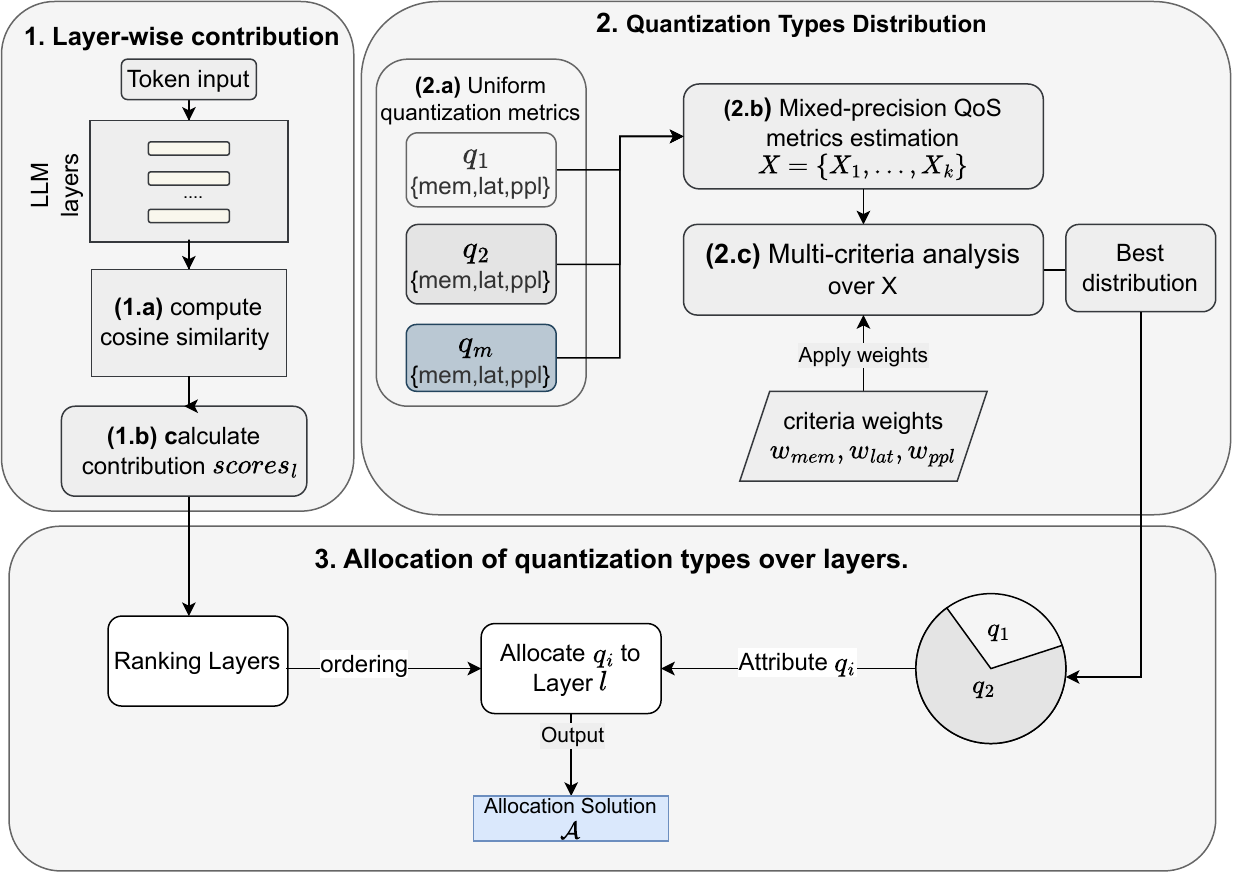}
    \caption{Contribution Overview}
    \label{fig:contrib-overview}
\end{figure}

We propose an adaptive mixed precision quantization solution based on three main modules:
(1) \textbf{Layer-wise contribution module} that calculates a score per layer according to its importance in the inference process~(Fig~\ref{fig:contrib-overview}.1). Specifically, it uses cosine similarity to measure how much the layer modifies the intermediate representation (i.e., hidden embedding). Layers that cause larger changes receive higher scores. (2)~\textbf{Quantization types distribution module}:
The objective of this module is to evaluate how many layers should be assigned a given quantization level by considering user preferences in terms of QoS metrics (provided as weights). It outputs the best distribution of quantization levels across the layers. To do so, we first need to explore all feasible solutions~(Fig.~\ref{fig:contrib-overview}.(2.b)), i.e., all possible ways to distribute the number of layers across quantization types based on uniform quantization measures (Fig.~\ref{fig:contrib-overview}.(2.a)). Then the solutions are ranked using the TOPSIS multi-criteria analysis method (Fig.~\ref{fig:contrib-overview}.(2.c)).
(3)~\textbf{Layer-wise quantization allocation}: this module maps a quantization level to each layer according to the scores of module (1) and respecting the quantization types distribution from module (2). This way, this module builds a mixed layer-wise quantization that takes into consideration the layer's importance along with the trade-offs between the three QoS metrics according to user preferences (Fig.~\ref{fig:contrib-overview}.3).

\subsection{Layer-wise contribution module~(Fig~\ref{fig:contrib-overview}.1)}
\label{layer-wise-contrib}
\label{lcm}
The core idea behind this module is to calculate a score per layer depending on its contribution, as in LLMs, the embeddings are progressively transformed as they pass through the model's layers. Some layers introduce substantial changes to the hidden embeddings and are more likely to play a critical role in the output.

To quantify these changes, we measured the cosine similarity between the hidden states of successive layers across a diverse set of prompts. The cosine similarity of the hidden embeddings of layers $i$ and $i+1$ is given as follows:
% \vspace{-11pt}

$$
    dist(i,i+1)=\frac{\mathbf{h}^{i}\cdot \mathbf{h}^{i+1}}{\|\mathbf{h}^{(i)}\| \, \|\mathbf{h}^{(i+1)}\|}
$$

We propose a Reward-Penalty approach that uses layer-wise statistics in order to estimate the layer's contribution. Each layer $i$ is evaluated by comparing its output to the next layer $i+1$; if the distance is lower than a fixed threshold $\gamma$, we increase its reward $R^{i}$, reflecting an information gain; otherwise, a penalty $P^{i}$ is increased.
The value of $\gamma$ was set to 0.9 as a heuristic balance, high enough to capture meaningful differences, yet not so low that most layers are uniformly penalized.
The score is computed as the difference between its reward and penalty $(score_{i}= R^{i}-P^{i})$ to enable a fair comparison between layers.
Finally, the layers are ordered depending on their contribution score.

\subsection{Quantization types distribution (Fig~\ref{fig:contrib-overview}.2) } 
\label{dqt} 
This module determines the number of layers associated with each quantization type.
A key consideration here is that each uniform quantization type has distinct QoS metrics in terms of memory, accuracy, and latency~(see Section~\ref{motiv:inf-lat}). 
This module exploits the complementary strengths of mixed quantization. To this goal, different layers in a model can benefit from different quantization types. Indeed, assigning a quantization type that offers low latency to more layers will reduce the overall inference latency; the same applies to the other metrics.

This module is built upon two parts. The first one devises a method to estimate the QoS metrics of all the feasible solutions of mixed quantizations, ie, distributions of layers across quantization types. 
Then, in a second part, we apply multi-criteria analysis to rank candidates and identify the best configuration.

\subsubsection{Mixed-precision QoS metrics estimation}

In order to evaluate the performance of a given mixed quantization solution, the ideal approach is to evaluate its QoS metrics through real measurements. For instance, for the Llama3.1 model, this would require measuring over 814385 models when using 5 types of quantization levels and 64 layers (that would take $\sim$42 years on a 30mn per model basis). Because of this combinatorial search space explosion,  we relied on estimation rather than measurements. We used the QoS metric values associated with uniform quantization types as building blocks to derive the QoS metrics of a mixed solution.

Formally, let I be the set of $M$ different uniform quantization types
$I= \{q_{i}, i \in \{ 1,.,M \} \}$.

A candidate solution $k$ is one way of dividing the $L$ layers among the $M$ quantization types, represented by a vector $
Z_k = (z_{k1}, \dots, z_{kM})$, where
$z_{ki}$ denotes the number of layers using type $q_i$; with  $\sum_{i=1}^M z_{ki}=L$.

As stated above, evaluating all possible solutions by directly measuring their QoS metric (e.g., memory or latency) is infeasible; therefore, we approximate those solution utilizing the measurements of QoS metrics of uniform quantization in order to compare them. To this end, we put forth the simplified hypothesis that each layer of the model affects the $J$ QoS metrics (memory, latency, accuracy) in the same way. For example, if we have 10 layers and the total latency is 100ms, each layer is assumed to contribute by 10ms.
Based on this assumption, for every quantization type $q_{i}$ in $I$, we record a set of 
$J$ metric values (latency, memory footprint, and accuracy in our case), let $c_{ij}$ be the value of $j^{th}$ metric for the quantization type $q_{i}$.
Thus, the set of measurements can be expressed by matrix $C \in \mathbb{R}^{M \times J}$\label{measuredC-matrix}.

The metric vector $X_k = (x_{k1}, \dots, x_{kJ})$  of each candidate solution $Z_k$ is approximated by a weighted combination of the uniform-type QoS metrics ($C \in \mathbb{R}^{M \times J}$), with weights given by the assignment proportions $\frac{z_{ki}}{L}$:
%\vspace{-0.5cm}

$$x_{kj}=\sum_{i=1}^{I} \frac{z_{ki}}{L}\cdot c_{ij} $$

Here, $x_{kj}$ is the estimated value of the QoS metric $j$ for solution $k$, reflecting the mixture of quantization types according to their layer proportions.
The output of this part is the estimation of $K$ distribution solutions ($Z$) over the $J$ metrics, which we denote by $X \in \mathbb{R}^{K\times J}$.

\subsubsection{ Multi-criteria Analysis  }
The approximated  QoS metrics vectors of the candidate solutions ($X_{k} \in \mathbb{R}^{J}$), are then compared through a multi-criteria analysis\footnote{In multi-criteria analysis (MCDA), metrics are typically referred to as criteria}, which take into account the weight attributed to each metric $w_j \in [0,1] $ , $W=\{w_{j}, j \in \{1,..,J\}\}$.
For our contribution, J=3, with $w1$, $w2$, $w3$ denoting the memory, latency, and accuracy QoS metrics, respectively.

These weights let the user emphasize different goals depending on the scenario, such as prioritizing latency in real-time applications or memory in resource-constrained devices.
For example, if latency is prioritized, the weight vectors can be \{$w_1$=0.1,$w_2$=0.8,$w_3$=0.1\}.

We then apply the TOPSIS method, which is highly regarded for solving the multiple attribute decision making \cite{chakraborty2022topsis}.
TOPSIS assumes that the best option is the one nearest to the ideal solution and farthest from the worst solution.

Firstly, the metric values $X_k$ of the solution $k$ are normalized by the Euclidean norm following this equation:

$y_{kj}=\frac{x_{kj}}{\sqrt{\sum_{k=1}^{K} x_{kj}^{2}}}$

Secondly, the importance of each metric is incorporated by multiplying the normalized value $y_{kj}$ by the weight value related to that metric $w_j$, therefore $a_{kj}=y_{kj} \cdot w_j$.
We denote $A_{k}$ as the normalized weighted metric vector of solution $k$.
Next, the ideal and negative-ideal solutions are defined. The ideal solution $A^*=\{a^{*}_{j}\}_{j=1}^{J}$ corresponds to the best attainable values across the $J$ QoS metrics, while the negative-ideal solution $ A^{-}=\{a^{-}_{j}\}_{j=1}^{J} $ corresponds to the worst values.

Finally, the ranking score of solution $k$ is computed by

$ranking\_score_k=  \frac{\sqrt{\sum_{j=1}^{J} (a_{kj}-a^{-}_{j})^{2}}}{\sqrt{\sum_{j=1}^{J} (a_{kj}-a^{-}_{j})^{2}}+\sqrt{\sum_{j-1}^{J} (a_{kj}-a^{*}_{j})^{2}} }$

The highest ranking score indicates the best alternative solution; therefore, we get the best candidate solution for the distribution of quantization types. 

Note that APreQEL does not exclude uniform solutions; if a uniform configuration is optimal under the specified weighting of QoS metrics, it will be selected. APreQEL goes further by integrating user-defined priorities and supporting mixed quantization schemes.

\subsection{Layer-wise quantization allocation (Fig~\ref{fig:contrib-overview}.3) }

In this part, the most contributing layers are mapped to the least aggressive quantization types, while less critical layers are progressively assigned more aggressive ones.
This is done by combining the ranking of layers using the contribution score (Section~\ref{lcm} ) with the distribution strategy (Section~\ref{dqt}) to assign the most suitable quantization type to each layer.

The Algorithm~\ref{algo:allocation} sorts quantization types from least to most aggressive and ranks layers by their contribution scores, from the most important to the least important (line 5). It iterates over the ordered quantization set $I$, then uses the best distribution $Z_{best}$ to determine the number of layers to associate with each quantization type (inner for loop)(lines 6-9). The algorithm maps higher-ranked layers to less aggressive quantization and lower-ranked layers to more aggressive ones, yielding the final allocation $\mathcal{A}$ (lines 10–14).

This integration ensures that layer contributions and the overall performance trade-offs according to the user's priorities are jointly respected in the quantization design.

\begin{algorithm}[H]
\caption{Layer-wise Quantization Allocation }
\begin{algorithmic}[1]
\footnotesize
\State \textbf{Inputs:} layer's contribution scores vector $Scores = \{score_1, \dots, score_L\}$; \\
    Best distribution $\mathbf{Z_{best}} = (z_1, \dots, z_M)$  \\Quantization types $I = \{q_1, \dots, q_M\}$
\Comment{ordered from least to most aggressive}
    \State \textbf{Output:} Allocation $ \mathcal{A}= \{ \mathbf{v}_1, \dots, \mathbf{v}_L \}$

        \State $order \gets$ indices of $Scores$ sorted in descending order
   \State $p \gets 0$

    \For{$i = 1$ to $M$} \Comment{iterate over quantization types}
        \For{$j = 1$ to $z_i$} \Comment{assign $z_i$ layers to type $q_i$}
                    \State $\ell \gets order[p]$ \Comment{layer index with $p$-th largest score}

            \State $\mathcal{A}[\ell] \gets q_i$
            \State $p \gets p+1$
        \EndFor
    \EndFor \\
    \Return $\mathcal{A}$ 
\end{algorithmic}
\label{algo:allocation}

\end{algorithm}

\section{Evaluation and Discussion}

% le plan :

\label{eval-sect}

\subsection{Experimental Methodology} \label{experiment_method}

We evaluate APreQEL across three different model architectures (Phi3.5, Llama3.1 and Qwen3-4B) against the uniform quantization types. In effect, it was not feasible to compare with past work, as none considers all three metrics jointly. Specifically, 
we aim to demonstrate that by leveraging mixed quantization, APreQEL can adjust to varied QoS metrics weightings while ensuring that the resulting models lie on the Pareto front.

\subsubsection{Experimental setup }\label{weight-category}

We used the K-type quantization types of llama.cpp for 3-, 4-, 5-, and 6-bit precision, and q8\_0 for 8-bit, for the uniform baseline model (except for Qwen3-4B where the 8-bit precision was skipped as it was dominated by the 6-bit one).

To evaluate APreQEL, we generated a set of 28 models. To do so, we have tuned the weights assigned to the three QoS metrics (memory, perplexity, and latency) in a way to span a large configuration space, as described below.

We divided the set of generated models into four categories: \textbf{APreQEL-Fairness}, where weights were distributed evenly across two or three metrics among 4 models \{($\frac{1}{3},\frac{1}{3},\frac{1}{3}$), (0.5,0.5,0), (0.5,0,0.5), (0,0.5,0.5)\}; \textbf{APreQEL-Latency} category, where latency is prioritized; and likewise for \textbf{APreQEL-Accuracy} and \textbf{APreQEL-Memory}.

The three latter APreQEL-categories prioritize a single metric across two scenarios. The first spans from moderately to highly skewed or dominant preferences across the three metrics; the prioritized one is given weights of 0.7, 0.8, and 0.9 (giving 3 models per category, totaling 9), while the other two metrics share the residual weight equally (e.g., $(0.8,0.1,0.1)$), and dominant where only one metric matters (e.g,$(1,0,0)$), giving 3 additional models. 
The second emphasizes only two metrics, using skewed pairwise weights of 0.75 (e.g., $(0.75, 0.25,0)$, giving 6 models), and 0.9 (e.g.$(0.1,0.9,0)$, resulting in 6 models).

To evaluate the resulting models, we measured the perplexity using the wiki-test-2 dataset, and the benchmarking tool of llama.cpp with GPU execution enabled for the execution time. We also modified the llama.cpp library to generate the mixed precision models. 

\vspace{-0.2cm}
\subsubsection{Evaluation metrics}  We used the following metrics:

\hspace{-0.35cm}\textbf{(1) Inference latency:} time to generate one token (ms/token).

\hspace{-0.35cm}\textbf{(2) Memory footprint of the model:} measured by the memory space occupied by the model parameters.

\hspace{-0.35cm}\textbf{(3) Perplexity:} assess the model's ability to generate cohesive and fluent sentences; lower perplexity corresponds to better predictive ability. It serves as an accuracy proxy for quantized LLMs~\cite{jin2024comprehensive}, as it is negatively correlated with accuracy.

    \hspace{-0.35cm}\textbf{(4) Hypervolume gain:} 
    To compare our solution set with the baseline uniform set across the three objectives, we relied on the hypervolume (HV) indicator, a set-quality indicator over multi-objective optimizations, which evaluates the quality of a set of solution~\cite{10.1145/3453474}.
    It measures the portion of the objective space dominated by the solutions relative to a reference point.

\begin{figure}[t]
        \centering
        \includegraphics[width=0.5\linewidth]{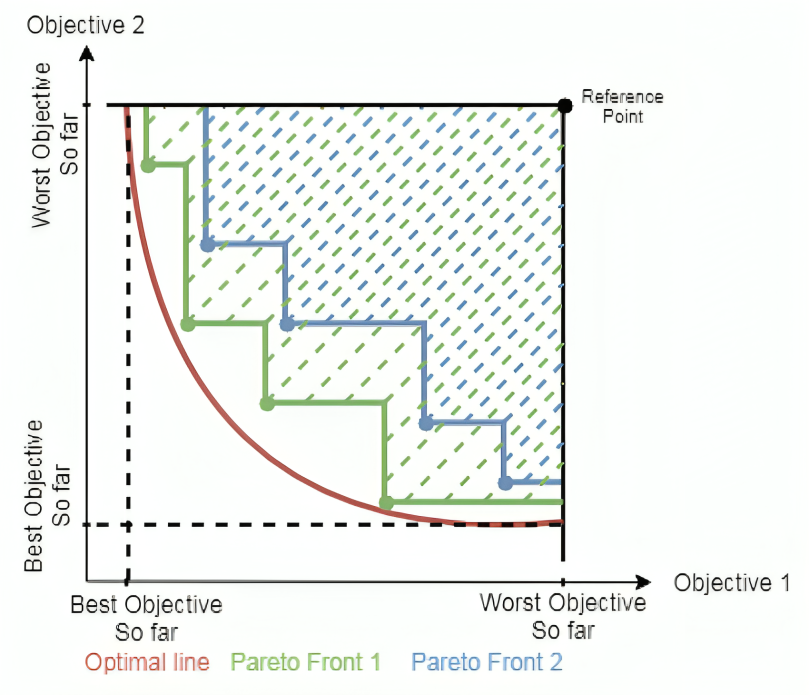}
        \caption{Hypervolume indicator\cite{8756240}}
       
        \label{fig:hyperV}
    \end{figure}

    Figure~\ref{fig:hyperV} illustrates how the hypervolume is calculated using the two objectives; it highlights the region that the solutions cover between the reference point and the optimal Pareto front (optimal line in red in the Figure). The closer the uncovered Pareto front is to the optimal one, the largest $HV$ is.
    The green Pareto front has a higher $HV$ as it covers more space closer to the optimal line, while the space between the green and the blue front represents the gain of the former over the latter.
    
    We note the $HV_{uni}$ as the hypervolume of the uniform set and $HV_{APreQEL}$ as the hypervolume associated with our APreQEL solutions, we define the gain as follows:
    $HV_{gain}=\frac{HV_{APreQEL}-HV_{uni}}{HV_{uni}}$

\subsubsection{Hardware setup}\label{hardware-setup}
We used the Jetson Orin-AGX embedded board, a 12-core Arm® v8.2 CPU, NVIDIA Ampere GPU architecture, and 64 GB LPDDR5 memory.

\subsection{Discussion}

We first discuss how APreQEL performs better than uniform quantization. Then, we evaluate how APreQEL generated models are effectively driven by users' predefined weights. Finally, we explore the effective choice made by APreQEL in terms of hybrid quantization levels.

\textbf{APreQEL Vs uniform Trade-offs:} To explore how APreQEL provides superior trade-offs, we compared the HV of the 28 models produced over the different weight configurations mentioned above~(Section \ref{weight-category}), with the uniform quantization baseline set across the three evaluation metrics.
 As usually done in multi-objective optimization, the HV values were computed\footnote{ emoa R package \url{https://cran.r-project.org/web/packages/emoa/index.html}} with respect to the worst metric values observed under the uniform quantization baseline, which served as the reference point.

 \begin{table}[h]
\centering
\begin{tabular}{|c|c|c|c|}
\hline
Model & $HV_{uni}$ & $HV_{APreQEL}$ & $HV_{gain}$ \\ \hline
Llama3.1 & 0.681 & 0.738 & 8.43\% \\ \hline
Phi3.5   & 0.247 & 0.270 & 9.07\% \\ \hline
Qwen3-4B  &  0.163 & 0.178 & 9.31\% \\ \hline
\end{tabular}
\caption{Comparison of Uniform and Our Method with HV gains for Llama3.1, Phi3.5 and Qwen3-4B}
\label{tab:hv_results}
\end{table}

\vspace{-0.7cm}
    Table \ref{tab:hv_results} reports the HV values for Llama3.1, Phi3.5, and Qwen3-4B for the uniform quantization baseline set $HV_{uni}$ and the APreQEL resulting models set $HV_{APreQEL}$. 

    APreQEL yields an $HV_{gain}$ of (8.43\%), (9.07\%), and (9.31\%) for the Llama3.1, Phi3.5, and Qwen3-4B models, respectively. Note that adding more configurations to APreQEL could further improve the hypervolume.
    These improvements show that applying mixed quantization with respect to user-defined QoS priorities uncovers new trade-offs that uniform quantization types cannot provide.
    
    A higher HV, as obtained by APreQEL,  reflects two aspects: (i)~ more solutions extend the Pareto front outward, meaning that the solution pushes at least one of the three metrics toward a better value.
    (ii)~ solutions achieve broader coverage across the QoS metrics, capturing a wider variety of trade-offs as compared to uniform quantization types.
    APreQEL guaranties flexibility by offering a larger number of configurations and ensuring that every solution remains competitive (non-dominated) in at least one objective.

\begin{table}[h]
\centering
\small
\scalebox{0.9}{
\begin{tabular}{|c|c|c|c|c|}
\hline
\textbf{Model} & \textbf{Category} & \textbf{Mem (GB)$\downarrow$} & \textbf{Lat (ms)$\downarrow$} & \textbf{PPL$\downarrow$} \\
\hline
\multirow{5}{*}{\textbf{Phi3.5}} 
 & APreQEL-Fairness & 0.573 & 1.952 & 0.493 \\
 & APreQEL-Latency  & 0.487 & \color{ForestGreen}\textbf{0.413} & 0.556 \\
 & APreQEL-Memory   & \color{ForestGreen}\textbf{0.170} & 36.705 & 2.122 \\
 & APreQEL-Accuracy & 1.143 & 14.130 & \color{ForestGreen}\textbf{0.099} \\
 % & Uniform          & 1.006 & 25.234 & 0.696 \\ 
 \hline \hline
\multirow{5}{*}{\textbf{Llama3.1}}
 & APreQEL-Fairness & 1.897 & 3.728 & 0.348 \\
 & APreQEL-Latency  & 1.255 & \color{ForestGreen}\textbf{1.232} & 0.416 \\
 & APreQEL-Memory   & \color{ForestGreen}\textbf{0.385} & 66.345 & 1.163 \\
 & APreQEL-Accuracy & 3.298 & 23.462 & \color{ForestGreen}\textbf{0.052} \\
% & Uniform          & 2.120 & 39.917 & 0.505 \\
\hline \hline
\multirow{5}{*}{\textbf{Qwen3-4B}}
 & APreQEL-Fairness & 0.610 & 2.270 & 0.740 \\
 & APreQEL-Latency  & 1.255 & \color{ForestGreen}\textbf{0.520} & 0.416 \\
 & APreQEL-Memory   & \color{ForestGreen}\textbf{0.100} & 51.070 & 2.890 \\
 & APreQEL-Accuracy & 1.240 & 23.070 & \color{ForestGreen}\textbf{0.148} \\
% & Uniform          & 0.730 & 26.640 & 1.130 \\
\hline
\end{tabular}}
\caption{Category-average difference to best metrics (lower is better) for Phi-3.5, LLaMA-3.1, and Qwen-3-4B.}
% \mb{is this table ok? is gap better than distance?}
% \mb{gaps} \jb{difference ?}\mb{ok}
\label{tab:3models-categories}
\end{table}

% ----phi3 on vertical first column

% \begin{table}[h]
% \centering
% \begin{tabular}{|c|c|c|c|c|}
% \hline
% \multirow{5}{*}[-2em]{\centering\rotatebox{90}{PHI3}}
% & Category & Memory(GB) $\downarrow$ & Latency (ms) $\downarrow$ & Perplexity $\downarrow$ \\ \cline{2-5}

% & APreQEL-Fairness & 0.573 & 1.952 & 0.493 \\ \cline{2-5}
% & APreQEL-Latency  & 0.487 & \color{ForestGreen}\textbf{0.413} & 0.556 \\ \cline{2-5}
% & APreQEL-Memory   & \color{ForestGreen}\textbf{0.170} & 36.705 & 2.122 \\ \cline{2-5}
% & APreQEL-Accuracy & 1.143 & 14.130 & \color{ForestGreen}\textbf{0.099} \\ \cline{2-5}
% & Uniform          & 1.006 & 25.234 & 0.696 \\ \hline

% \end{tabular}
% \caption{Distance of category-average to the best evaluation metric of Phi3.5 (lower is better).}
% \label{tab:phi3.5_category}
% \end{table}
% $\bullet$~ 

\textbf{APreQEL's Robustness}

In this part, we investigate  APreQEL's ability to meet user QoS demands (fixed through weights). For instance, we want to ensure that if the user requests a model with a low memory footprint (with a weight higher than the others), the generated model effectively decreases this metric. To put it concisely, each model category should demonstrate the best results according to the emphasized metric. To do so, we have measured how far each category is from a given metric's best model. 
Also, for a given QoS metric, the best model should be one of the categories dedicated to that metric (that is, the best model for latency should be a model for which the user has given a higher weight for that metric).
The latter condition was satisfied in all our tests; that is, the best model for each QoS metric was one for which a high weight was given.

Table \ref{tab:3models-categories} reports the difference between each APreQEL category’s average and the best value of the corresponding QoS metric for Phi3.5, Llama3.1 and Qwen3-4B models. 
APreQEL-Memory achieves the closest results to the minimal memory footprint, with small differences (in green) of 0.17 (Phi-3.5), 0.385 (Llama-3.1), and 0.1 (Qwen3-4B). Similarly, APreQEL-Latency and APreQEL-Accuracy excel at minimizing the differences to best latency and best perplexity, respectively, for the three evaluated models. We also note that, while APreQEL-Memory effectively reduces memory use, it exhibits higher perplexity, whereas APreQEL-Fairness maintains a more balanced trade-off across the three QoS metrics and never yields the worst deviation for any metric.

 Moreover, APreQEL may use solely one quantization type for all layers to achieve the exact minimal value for the prioritized metric or combine multiple quantization types to balance several QoS objectives, depending on how the metrics are weighted.
 
 These findings demonstrate APreQEL's robustness in meeting specific QoS goals and generalizing across different model architectures. This behavior fits the needs of edge deployment, since operational stability depends on selecting the configuration that meets the practical requirements; shrinking the memory footprint for resource-limited devices, reducing latency for real-time applications, focusing on accuracy for critical tasks, or achieving a balanced QoS over the system.

\begin{figure}[ht]
\centering
    \includegraphics[width=1\linewidth]{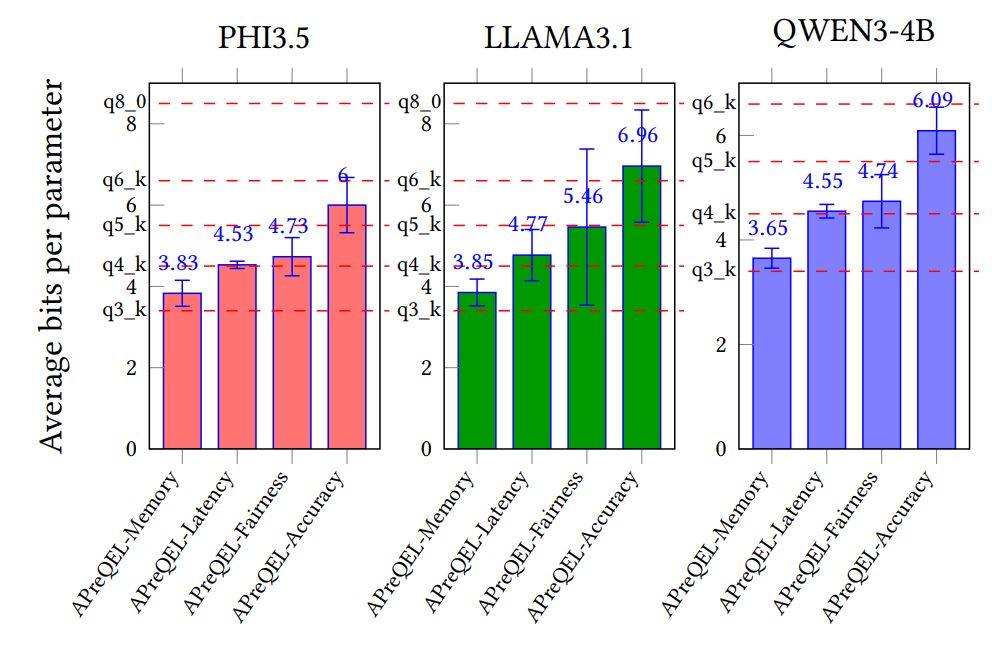}
 \caption{Average bit-width for APreQEL-categories for the three models}
\label{averge-bit-width}
\end{figure}

\textbf{APreQEL's adaptibility} In this part, we evaluate how APreQEL adapts the distribution of quantization types in response to different QoS priorities and model architectures. To capture this, we report, in Figure~\ref{averge-bit-width}, the average bit-width of the models produced for the APreQEL categories, across the three models' architectures, along with the variability of these bit-widths within each category.
The dashed red markers identify the fixed bit-widths of the uniform quantization baselines~(q3\_K,..,q8\_0).

We note that the effective bit-widths used by the uniform quantization exceed the nominal precision. This is because the parameters of the model are partitioned into groups, and each group is quantized with a scaling factor. This is why, for example, the q3\_K quantization has a bit-width of 3.44 instead of 3.

APreQEL is not constrained to a fixed bit-width; depending on the weight configuration, it can result in a uniform assignment or generate mixed-precision that allocates different quantization types to layers. This flexibility enables finer control over quantization behavior. Across all three models, the configurations line up in a consistent pattern: APreQEL-Memory settings always attribute the lowest bit-widths, between 3 and 4-bit precisions; APreQEL-Accuracy uses higher precision to preserve model quality, while APreQEL-Latency assigns 4-bit precision to most layers, as it performs best on latency among the three models (see Fig\ref{fig:cosine-sim-ffn-att}). APReQEL-Fairness produces a moderate bit-width pattern, aiming to balance the competing objectives.
The exact magnitudes and spreads, however, vary by model, indicating that each model architecture translates these optimization priorities into distinct distributions.

For example, under the weight setting (0.7, 0.1, 0.1), which places the highest emphasis on memory, APreQEL assigns 72.2\% of Qwen3-4B’s 72 layers to q3\_K and 27.7\% to q4\_K. In contrast, for Phi-3.5 (64 layers), 39\% of layers are mapped to q3\_K and 60.9\% to q4\_K, while for Llama-3.1 (64 layers), 42.18\% are assigned to q3\_K and 57.8\% to q4\_K, showing that the mixed-precision pattern is not constant but model-dependent.

Indeed, APReQEL explicitly accounts for the model-specific differences in QoS behavior across uniform quantization types (overview (Fig\ref{fig:contrib-overview}.(2.a)), allowing it to enable a systematic way to obtain, under weighting configuration, the best distribution of quantization types across different models. To the best of our knowledge, this aspect has not been studied in any prior work. Consequently, APreQEL adapts the quantization types not only to the chosen optimization priorities but also to the characteristics of each model architecture.

\section{Conclusion and Perspectives}
In conclusion, we introduced APreQEL, a novel framework that leverages mixed quantization to enable efficient LLM inference on resource-constrained edge devices by uncovering a large set of configurations optimized according to three QoS metrics; latency, memory footprint and accuracy.
APreQEL systematically assigns the most suitable quantization types to the appropriate layers under user-defined prioritization. This design makes it robust and adaptable to fluctuating edge environments.

For future work, we intend to extend APreQEL to include other metrics (e.g., energy, carbon footprint), support different hardware (e.g., TPUs, NPUs), and mix quantization types from different libraries, provided they are compatible, as quantization algorithms often vary and may exhibit distinct trade-off.

%% The next two lines define the bibliography style to be used, and
%% the bibliography file.
\bibliographystyle{ACM-Reference-Format}
\bibliography{mybib}
% \printbibliography

\end{document}